\documentclass{article}

\usepackage{microtype}
\usepackage{graphicx}
\usepackage{subcaption}
\usepackage{booktabs} 
\usepackage[dvipsnames]{xcolor}
\usepackage{caption}

\usepackage{amsmath,amsfonts,bm}

\def\eqref#1{equation~\ref{#1}}

\def\1{\bm{1}}

\def\ervd{{\textnormal{d}}}

\def\rmI{{\mathbf{I}}}

\DeclareMathAlphabet{\mathsfit}{\encodingdefault}{\sfdefault}{m}{sl}
\SetMathAlphabet{\mathsfit}{bold}{\encodingdefault}{\sfdefault}{bx}{n}

\def\gE{{\mathcal{E}}}

\def\gH{{\mathcal{H}}}

\def\gL{{\mathcal{L}}}

\def\gN{{\mathcal{N}}}

\usepackage{hyperref}

\usepackage[accepted]{icml2025}

\usepackage{amsmath}
\usepackage{amssymb}
\usepackage{mathtools}
\usepackage{amsthm}

\usepackage[capitalize,noabbrev]{cleveref}

\theoremstyle{plain}

\theoremstyle{definition}

\theoremstyle{remark}

\usepackage[textsize=tiny]{todonotes}

\icmltitlerunning{Denoising Hamiltonian Network for Physical Reasoning}

\begin{document}

\twocolumn[{
\renewcommand\twocolumn[1][]{#1}%
\icmltitle{Denoising Hamiltonian Network for Physical Reasoning}



\icmlsetsymbol{equal}{*}

\begin{icmlauthorlist}
\icmlauthor{Congyue Deng}{mit,stf,equal}
\icmlauthor{Brandon Y. Feng}{mit}
\icmlauthor{Cecilia Garraffo}{cfa}
\icmlauthor{Alan Garbarz}{conicetuba}
\icmlauthor{Robin Walters}{neu}\\
\icmlauthor{William T. Freeman}{mit}
\icmlauthor{Leonidas Guibas}{stf}
\icmlauthor{Kaiming He}{mit}
\end{icmlauthorlist}

\icmlaffiliation{mit}{Massachusetts Institute of Technology}
\icmlaffiliation{stf}{Stanford University}
\icmlaffiliation{cfa}{Harvard-Smithsonian Center for Astrophysics}
\icmlaffiliation{neu}{Northeastern University}
\icmlaffiliation{conicetuba}{Universidad de Buenos Aires and Instituto de Física de Buenos Aires -- CONICET}

\icmlcorrespondingauthor{Congyue Deng}{congyue@stanford.edu}

\icmlkeywords{Physical learning, Physical reasoning, Hamiltonian Neural Network, Masked modeling, Denoising}


\vskip 0.1in

\begin{center}
    \centering
    \includegraphics[width=.9\linewidth]{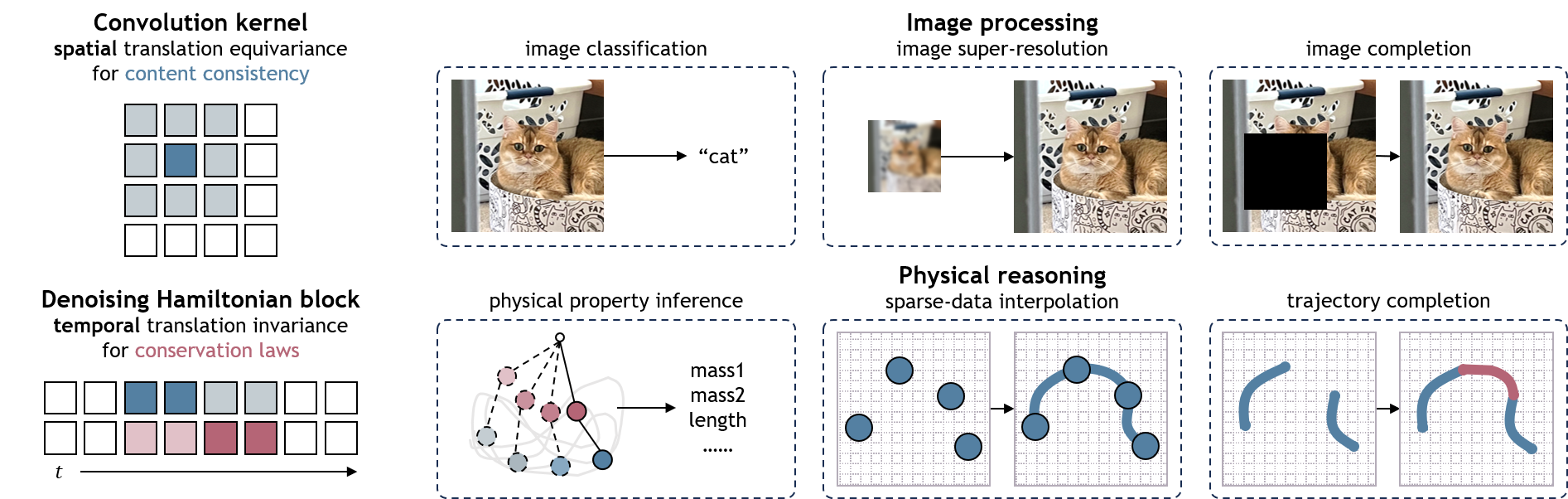}
    \vspace{-1em}
    \captionof{figure}{\textbf{Denoising Hamiltonian Network (DHN)} generalizes Hamiltonian mechanics into neural operators. It enforces physical constraints while leveraging the flexibility of neural networks, opening pathways for broader applications in physical reasoning.}
    \label{fig:teaser}
\end{center}

\vskip 0.1in
}]



\printAffiliationsAndNotice{\icmlEqualContribution} 

\begin{abstract}
Machine learning frameworks for physical problems must capture and enforce physical constraints that preserve the structure of dynamical systems.
Many existing approaches achieve this by integrating physical operators into neural networks.
While these methods offer theoretical guarantees, they face two key limitations: (i) they primarily model local relations between adjacent time steps, overlooking longer-range or higher-level physical interactions, and (ii) they focus on forward simulation while neglecting broader physical reasoning tasks.
We propose the Denoising Hamiltonian Network (DHN), a novel framework that generalizes Hamiltonian mechanics operators into more flexible neural operators. 
DHN captures non-local temporal relationships and mitigates numerical integration errors through a denoising mechanism.
DHN also supports multi-system modeling with a global conditioning mechanism.
We demonstrate its effectiveness and flexibility across three diverse physical reasoning tasks with distinct inputs and outputs.
\end{abstract}
\vspace{-2em}
\section{Introduction}

Physical reasoning -- the ability to infer, predict, and interpret the behavior of dynamic systems -- is fundamental to scientific inquiry.
Machine learning frameworks designed to address such challenges are often expected to go beyond merely memorizing data distributions, aiming to uphold the laws of physics, account for energy and force relationships, and incorporate structured inductive biases that surpass those of purely data-driven models.
Scientific machine learning addresses this challenge by embedding physical constraints directly into neural network architectures, often through explicitly constructed physical operators.

However, these methods face two key limitations.
(i) These methods primarily learn local temporal updates—predicting state transitions from one time step to the next—without capturing long-range dependencies or abstract system-level interactions.
(ii) They focus predominantly on forward simulation, forecasting a system's evolution from initial conditions, while largely overlooking complementary tasks such as super-resolution, trajectory inpainting, or parameter estimation from sparse observations.

To address these limitations, we introduce the \textbf{Denoising Hamiltonian Network (DHN)}, a framework that generalizes Hamiltonian mechanics into neural operators.
DHN enforces physical constraints while leveraging the flexibility of neural networks, leading to three key innovations.

First, DHN extends Hamiltonian neural operators to capture non-local temporal relationships by treating groups of system states as tokens, allowing it to reason holistically about system dynamics rather than in isolated steps.

Second, DHN integrates a denoising objective, inspired by denoising diffusion models, to mitigate numerical integration errors. 
By iteratively refining its predictions toward physically valid trajectories, DHN enhances stability in long-term forecasting while remaining adaptable across diverse noise conditions. 
Additionally, by leveraging different noise patterns, DHN supports flexible training and inference across various task contexts.

Third, we introduce global conditioning to facilitate multi-system modeling. A shared global latent code encodes system-specific properties (e.g., mass, pendulum length), enabling DHN to model heterogeneous physical systems under a unified framework while maintaining disentangled representations of underlying dynamics.

To evaluate DHN’s versatility, we test it across three distinct reasoning tasks: (i) trajectory prediction and completion, (ii) inferring physical parameters from partial observations, and (iii) interpolating sparse trajectories via progressive super-resolution.

In summary, this work moves toward more general network architectures that embed physical constraints beyond local temporal relationships, opening pathways for broader applications in physical reasoning beyond conventional forward simulation and next-state prediction.

\vspace{-.5em}
\section{Related Work}

Machine learning approaches for physical modeling span fundamental equations of motion to high-dimensional operator learning. Our work extends Hamiltonian neural networks (HNNs) into a flexible, sequence-based paradigm that enables multi-task inference and generative conditioning.

\vspace{-.5em}

\paragraph{Hamiltonian Neural Networks (HNNs)}
Scientific machine learning aims to embed physical laws into neural architectures. 
Hamiltonian Neural Networks (HNNs) \citep{greydanus2019hamiltonian} enforce symplectic structure and energy conservation in learned dynamics, inspiring various extensions: Lagrangian Neural Networks (LNNs) \citep{cranmer2020lagrangian}, Symplectic ODE-Nets \citep{zhong2020symplectic}, and Dissipative SymODEN \citep{zhong2020dissipative}, which introduce damping terms. 
Constraints have also been incorporated into HNNs \citep{finzi2020simplifying}, and some models infer Hamiltonian dynamics directly from image sequences \citep{toth2020hamiltonian}.
Despite their strengths in forward simulation, standard HNNs typically model one system at a time and rely on uniform-step integration, limiting their use in trajectory completion, sparse-data interpolation, or super-resolution.

\vspace{-.5em}

\paragraph{Physics-informed and operator-based methods}
Another approach embeds partial differential equation (PDE) constraints directly into neural models. Physics-Informed Neural Networks (PINNs) \citep{raissi2019physics} enforce PDE-based losses for solving forward and inverse problems, while Fourier Neural Operators (FNOs) \citep{li2021fourier} learn mappings between function spaces using global Fourier transforms. Neural ODEs \citep{chen2018neural, dupont2019augmented} parameterize continuous-time dynamics with learnable differential equations. While these methods effectively model spatiotemporal PDEs, they are less suitable for discrete Hamiltonian dynamics with irregular sampling. In contrast, our method directly operates on discrete Hamiltonian structures using block-wise transformations, enhancing flexibility while preserving interpretability and stability.

\vspace{-.5em}

\paragraph{System identification and multi-system modeling}
Learning from heterogeneous physical systems requires system identification, traditionally performed via parametric models \citep{ljung1998system} or hybrid PDE-constrained approaches \citep{raissi2019physics}. While Hamiltonian methods implicitly encode system parameters through energy landscapes, conventional HNNs often require training separate models per system. We introduce a generative conditioning mechanism via a learned latent code, enabling a single model to generalize across multiple systems while preserving inductive biases from Hamiltonian dynamics.
\section{Method}

\subsection{Motivation}

\begin{figure}[t]
    \centering
    \includegraphics[width=.9\linewidth]{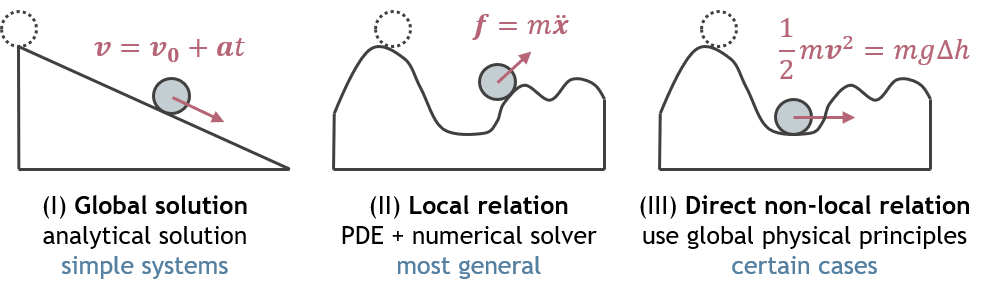}
    \vspace{-1em}
    \caption{\textbf{How can we solve for a physical state?}
    {(I)} Closed-form analytical solutions for simple systems.
    {(II)} For more complex physical systems, most physical PDEs only model local relations of close-by time steps.
    {(III)} For certain physical systems, states can be directly related even if they are not close by temporally.
    }
    \label{fig:motivation}
    \vspace{-1em}
\end{figure}

Our goal is to design more general neural operators that both follow physical constraints and unleash the flexibility and expressivity of neural networks as optimizable black-box functions. We start by asking the question: \textit{What ``physical relations'' can we model beyond next-state prediction?}

Figure \ref{fig:motivation} compares three classical approaches to modeling physical systems without machine learning:
Case (I): Global Analytical Solution. For simple systems with regular structures, one often derives a closed‐form solution directly.
Case (II): PDE + Numerical Integration. In more complex settings where no closed‐form solution exists, the standard practice is to formulate the system’s dynamics as a PDE and solve it step‐by‐step over time via numerical methods. This local integration approach underlies most physics‐constrained neural network designs, which encode the PDE operators into the network to ensure physical consistency at each step.
Case (III): Direct Global Relation. In some complex systems (e.g., purely conservative systems without dissipative forces), states that are temporally far apart can be related directly using global conservation laws (e.g., energy conservation).
This is akin to high‐school physics problems: one can compute an object’s velocity at a certain position from initial conditions alone, without solving for a full trajectory. While this is less general than PDE-based approaches, it suggests a promising avenue: leveraging global physical principles within a black-box neural network could extend this technique to more complex, real-world dynamical systems beyond simple textbook problems.

\subsection{Preliminaries}
\label{sec:preliminaries}

\begin{figure}[t]
    \centering
    \includegraphics[width=.7\linewidth]{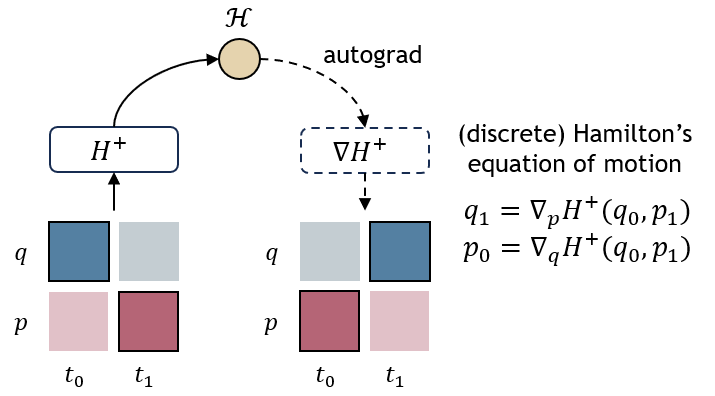}
    \vspace{-1.2em}
    \caption{\textbf{Discrete (right) Hamiltonian neural network.} Dark blue and dark red indicate network inputs and outputs. Light colors illustrate the adjacent time steps.}
    \label{fig:hnn}
    \vspace{-1em}
\end{figure}

\paragraph{Learning with Hamiltonian mechanics}
Let's start with \textit{phase-space coordinates} $(q, p)$, where $q$ is the \textit{generalized coordinates} and $p$ is the \textit{generalized momenta} or \textit{conjugate momenta}. 
If $q$ represents the particle positions in Euclidean coordinates, then $p$ corresponds to their linear momenta. If $q$ represents angular positions in spherical coordinates, $p$ corresponds to the associated angular momenta.
We consider the time-invariant \textit{Hamiltonian}, which is a scalar function $\gH(q,p)$ satisfying
\begin{equation}
\label{eq:eom}
    \frac{\ervd q}{\ervd t} = \nabla_p \gH, \quad
    \frac{\ervd p}{\ervd t} = -\nabla_q \gH.
\end{equation}
Eq. \ref{eq:eom} is known as Hamilton's equations of motion and describes system evolution by defining a trajectory in phase space along the vector field $(\nabla_p \gH, -\nabla_q \gH)$. This field, called the \textit{symplectic gradient}, governs the dynamics such that movement along $\gH$ induces the most rapid change in the Hamiltonian, whereas motion in the symplectic direction preserves the system's energy structure.

Hamiltonian Neural Networks (HNN)~\citep{greydanus2019hamiltonian} treat the Hamiltonian as a black-box function $\gH(q, p;\theta)$ parameterized by a neural network and optimize the network parameters to minimize the loss function
\begin{equation}
\label{eq:loss_hnn}
    \gL_{\text{HNN}}(\theta) =
    \left\| \nabla_p \gH - \frac{\ervd q}{\ervd t} \right\| +
    \left\| \nabla_q \gH + \frac{\ervd p}{\ervd t} \right\|.
\end{equation}
Starting with an initial state $(q_0, p_0)$, one can compute the trajectory $(q_t, p_t)$ by integrating the symplectic gradient $(\nabla_p \gH(q_t, p_t;\theta), -\nabla_q \gH(q_t, p_t;\theta)$ over time $t$.

\paragraph{Discrete Hamiltonian}
Aside from the continuous Hamiltonian $\gH$ and its discretizations, one can also directly define the discrete Hamiltonian with discrete mechanics and duality theory in convex optimization \cite{gonzalez1996time}. The discrete ``right'' Hamiltonian $H^+$ gives the equation of motion in the form
\begin{align}
    q_{t+1} &= \nabla_p H^+(q_t, p_{t+1}), \label{eq:h_plus_q} \\
    p_t &= \nabla_q H^+(q_t, p_{t+1}). \label{eq:h_plus_p}
\end{align}
The ``right'' means that $q$ is forward and $p$ is backward in time.
This formulation serves as a first-order discrete approximation of the continuous Hamiltonian $\gH$ by
\begin{align}
    q_{t+1} &= q_t + \Delta t \nabla_p \gH(q_t, p_{t+1}), \label{eq:h_plus_continuous_q} \\
    p_t &= p_{t+1} + \Delta t \nabla_q \gH(q_t, p_{t+1}). \label{eq:h_plus_continuous_p}
\end{align}
Figure \ref{fig:hnn} illustrates a discrete right Hamiltonian network for computing the state relations between time steps $t_0$ and $t_1$.
We describe our network design primarily using the right Hamiltonian $H^+$, but similar equations can define the left Hamiltonian $H^-$ and the same approach applies to $H^-$. Additional details can be found in Appendix \ref{sec:appen:h_minus}.

Exemplified by HNN, physical networks generally learn the state relations between adjacent time steps $t$ and $t+1$ modeled by an update rule
\begin{equation}
    (q_{t+1}, p_{t+1}) = \texttt{update\_rule}(q_t, p_t).
\end{equation}
Compared to forward modeling, the discretization in Eqs.~\ref{eq:h_plus_q} and \ref{eq:h_plus_p} is more accurate and better preserves the symplectic structure of the system under temporal integrations.
However, the implicit nature of these update rules introduces challenges at inference time, as determining new system states requires solving an optimization problem, which becomes difficult when the available data consists of a single simulation trajectory without additional reference points.

Our solution is to incorporate the optimization process into the network, leading to the \emph{denoising} Hamiltonian network (Sec. \ref{sec:denoising_hnn}) that unifies the denoising update rules for state optimization at each time step and the Hamiltonian-modeled state relations across time steps.

\subsection{Block-Wise Discrete Hamiltonian}
\label{sec:blockwise_hnn}

\begin{figure}[t]
    \centering
    \includegraphics[width=.95\linewidth]{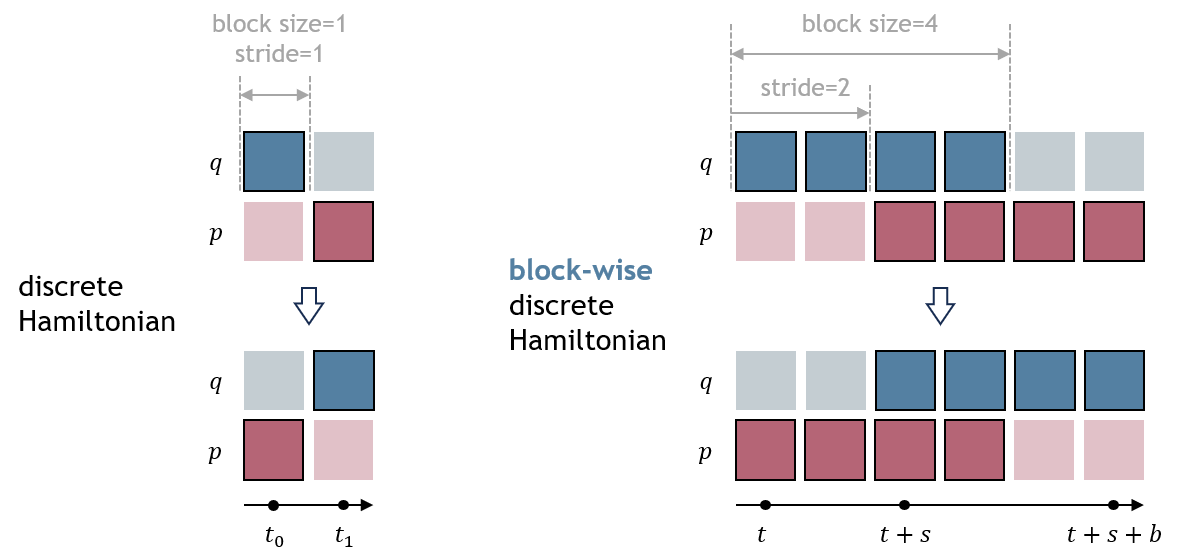}
    \vspace{-1em}
    \caption{\textbf{Block-wise Hamiltonian.}
    Left: Classical HNN viewed as a special case of block size $b=1$ and stride $s=1$.
    Right: A discrete (right) Hamiltonian block with $b=4, s=2$.
    Dark blue and dark red indicate network inputs and outputs. Light colors illustrate the adjacent time steps.
    }
    \label{fig:blockwise_hnn}
\end{figure}

We define state blocks as a stack of $(q, p)$ states concatenated along the time dimension $Q_{t}^{t+b} = [q_t, \cdots, q_{t+b}], P_{t}^{t+b} = [p_t, \cdots, p_{t+b}]$, with $b$ being the block size.
We also introduce the stride $s$ as a hyperparameter that can be flexibly defined, replacing the fixed time interval $\Delta t$ in Eqs.~\ref{eq:h_plus_continuous_q}-\ref{eq:h_plus_continuous_p}. 
This approach enables the network to capture broader temporal correlations while preserving the underlying Hamiltonian structure.
We define our block-wise discrete (right) Hamiltonian by relating two overlapping blocks of system states, each of size $b$ with a shift stride of $s$
\begin{align}
    Q_{t+s}^{t+s+b} &= \nabla_P H^+(Q_{t}^{t+b}, P_{t+s}^{t+s+b}), \label{eq:block_h_plus_q} \\
    P_{t}^{t+b} &= \nabla_Q H^+(Q_{t}^{t+b}, P_{t+s}^{t+s+b}).  \label{eq:block_h_plus_p}
\end{align}
Figure \ref{fig:blockwise_hnn} illustrates a block-wise discrete Hamiltonian of a block size $b=4$ and a stride $s=2$. Classical HNNs can be viewed as a special case of block size $b=1$ and stride $s=1$. Physical interpretations of the block-wise Hamiltonian with $b>1, s>1$ can be found in Appendix \ref{sec:appen:block_hnn}.

Similar to HNN, a block-wise discrete Hamiltonian network $H^+_\theta$ can be trained with the equation-of-motion loss following Eq. \ref{eq:block_h_plus_q}-\ref{eq:block_h_plus_p}
\begin{align}
\begin{split}
\label{eq:loss_dhn}
    \gL_{\text{block}}(\theta) =&~
    \left\| \nabla_P H^+_\theta(Q_{t}^{t+b}, P_{t+s}^{t+s+b}) - Q_{t+s}^{t+s+b} \right\| \\
    &+ \left\| \nabla_Q H^+_\theta(Q_{t}^{t+b}, P_{t+s}^{t+s+b}) - P_{t}^{t+b} \right\|.
\end{split}
\end{align}

\subsection{Denoising Hamiltonian Network}
\label{sec:denoising_hnn}

\begin{figure}[t]
    \centering
    \includegraphics[width=.9\linewidth]{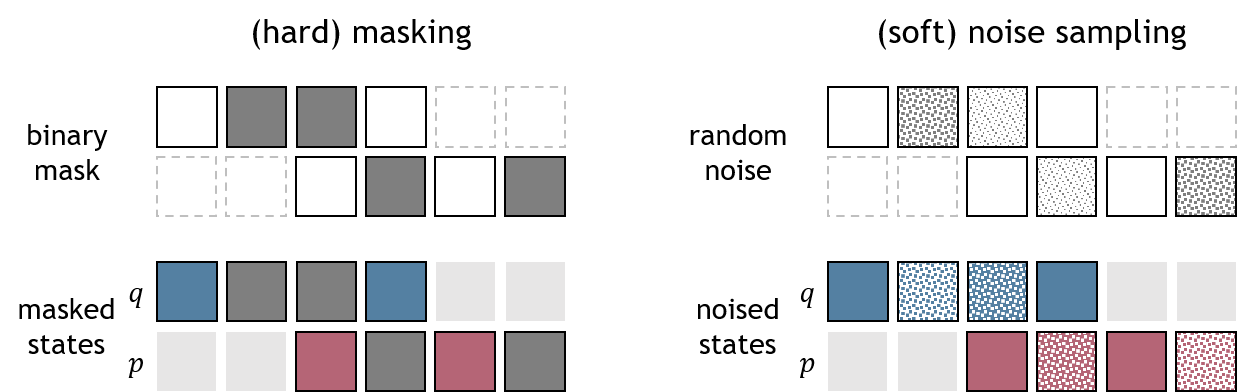}
    \vspace{-.5em}
    \caption{\textbf{Denoising Hamiltonian block.}
    Left: Random masking on input states.
    Right: Random noise sampling on input states. Different states have different sampled noise scales.
    }
    \label{fig:denoising_hnn}
\end{figure}

\paragraph{Masked modeling and denoising}
Following our motivations introduced in Sec. \ref{sec:preliminaries}, we want the Hamiltonian blocks to not only model the state relations across time steps, but also learn the state optimization per time step for inference. To achieve that, we adopt a masked modeling strategy \cite{he2022masked} by training the network with a part of the input states masked out (Figure \ref{fig:denoising_hnn}).

Rather than simply masking out input states, we perturb them with noise sampled at varying magnitudes (Figure \ref{fig:denoising_hnn}).
This strategy ensures that the model learns to refine predictions iteratively, enabling it to recover physically meaningful states from corrupted or incomplete observations.
Concretely, we define a sequence of increasing noise levels $0=\alpha_0 < \alpha_1 < \cdots < \alpha_N=1$.
Taking the blocked input state $Q_{t}^{t+b}$ as an example, we randomly sample Gaussian noises $\gE_{t}^{t+b} = [\varepsilon_t, \cdots, \varepsilon_{t+b}]$ and per-state noise scales $A_{t}^{t+b} = [\alpha_t,\cdots, \alpha_{t+b}]$.
Let $M_{t}^{t+b} = [m_t,\cdots,m_{t+b}]$ be the binary masks with 0 for unknown states and 1 for known states, we obtain the noised input $\widetilde{Q}_{t}^{t+b}$ by
\begin{align}
    A' &= A\cdot(1-M), \\
    \widetilde{Q} &= (1-A')\cdot Q + A'\cdot\gE.
\end{align}
Intuitively, it enforces the known states to have a noise scale of 0.
The number of denoising steps is set to 10 in our experiments.
At inference time,  we progressively denoise the unknown states with a sequence of decreasing noise scales that are synchronized on all unknown states.
We apply both $H^+$ and $H^-$ to iteratively update $(Q_{t}^{t+b}, P_{t+s}^{t+s+b})$ and $(Q_{t+s}^{t+s+b}, P_{t}^{t+b})$.
More details are in Appendix \ref{sec:appen:denoising}.

\begin{figure}[t]
    \centering
    \includegraphics[width=\linewidth]{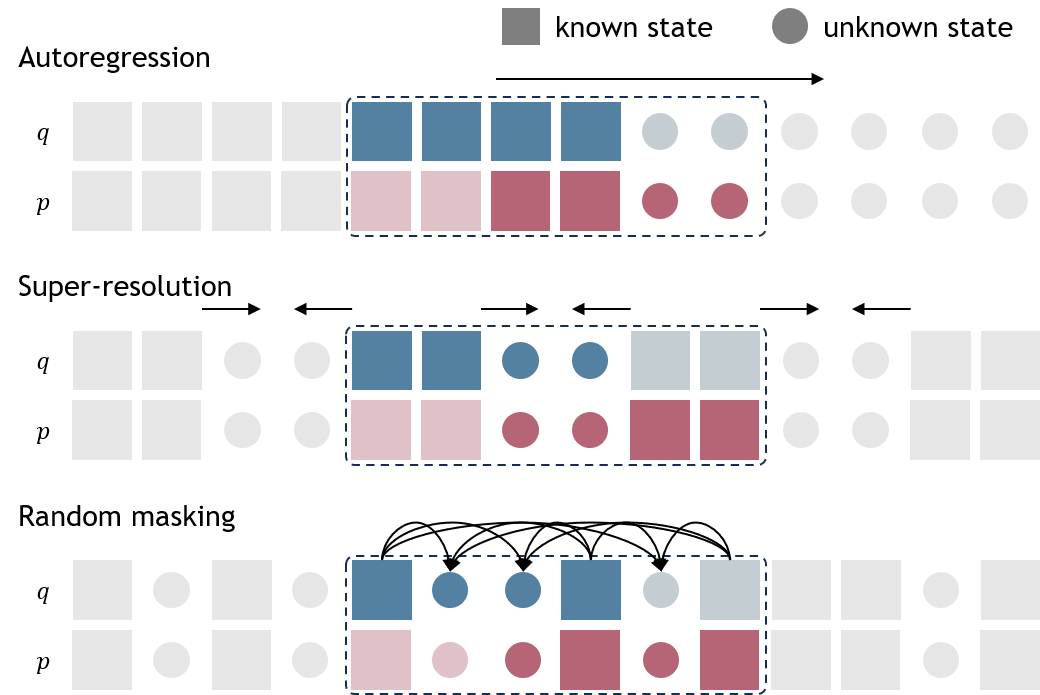}
    \vspace{-2em}
    \caption{\textbf{Different masking patterns.} Training with different masking patterns enables different inference strategies. Colored blocks surrounded by dotted lines are the denoising Hamiltonian blocks sliding along the sequences.}
    \label{fig:inference_types}
\end{figure}

\paragraph{Different masking patterns}
By designing distinct masking patterns during training, we enable flexible inference strategies tailored to different tasks.
Figure \ref{fig:inference_types} shows three types of different masking patterns: {\it autoregression} by masking out the last few states of a block, which resembles physical simulation in terms of next-state prediction with forward modeling; {\it super-resolution} by masking out the states in the middle of a block, which can be applied to data interpolation; and more generally, {\it arbitrary-order} masking including random masking, with the masking pattern adaptively designed according to the task requirements.

\subsection{Network Architecture}
\label{sec:network_architecture}

\begin{figure}[t]
    \centering
    \includegraphics[width=.9\linewidth]{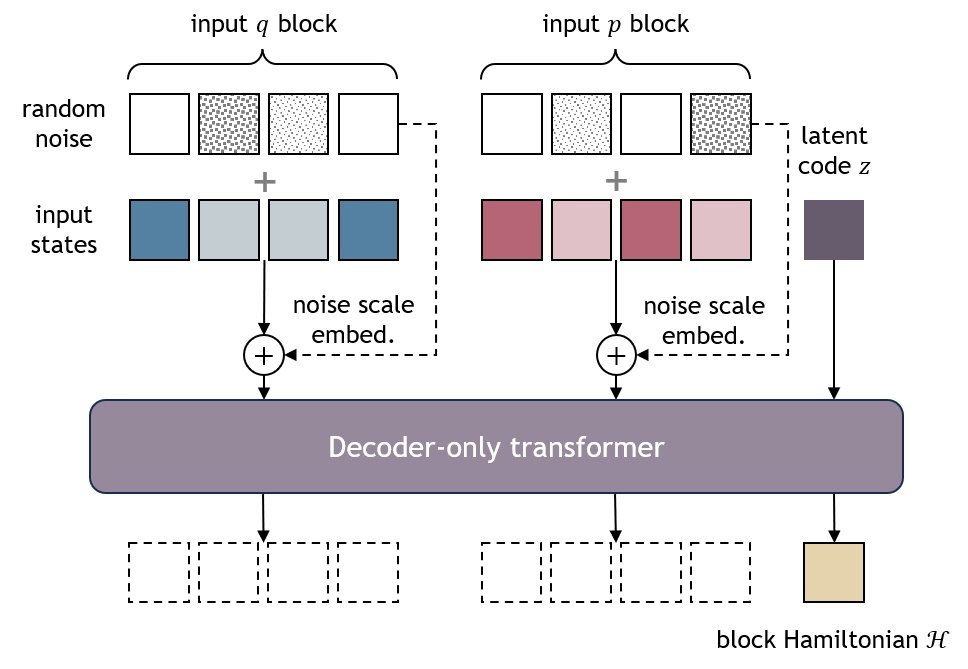}
    \vspace{-1.5em}
    \caption{\textbf{Decoder-only transformer architecture.}
    We use a latent code $z$ for each trajectory to serve as the query token for the Hamiltonian value output.
    Per-state noise scales are encoded and added to the positional embeddings.
    Dark purples (in all shades) indicate trainable modules or variables.}
    \label{fig:network_transformer}
\end{figure}

\paragraph{Decoder-only transformer}
For each Hamiltonian block, the network inputs are a stack of $Q_{t}^{t+b}$ of different time steps, a stack of $P_{t'}^{t'+b}$, and we also introduce a global latent code $z$ for the entire trajectory as conditioning. We employ a decoder-only transformer \cite{radford2019language, jin2024lvsm}, which resembles a GPT-like decoder-only architecture but without a causal attention mask, as shown in Figure \ref{fig:network_transformer}. We apply self-attention to all input tokens $[Q_{t}^{t+b}, P_{t'}^{t'+b}, z]$ as a sequence of length $2b+1$. The global latent code $z$ serves as a query token for outputing the Hamiltonian value $\gH$.
We also encode the per-state noise scales into the network by adding their embeddings to the positional embedding.
In our experiments, we implement a simple two-layer transformer that fits into a single GPU.

\begin{figure}[t]
    \centering
    \includegraphics[width=\linewidth]{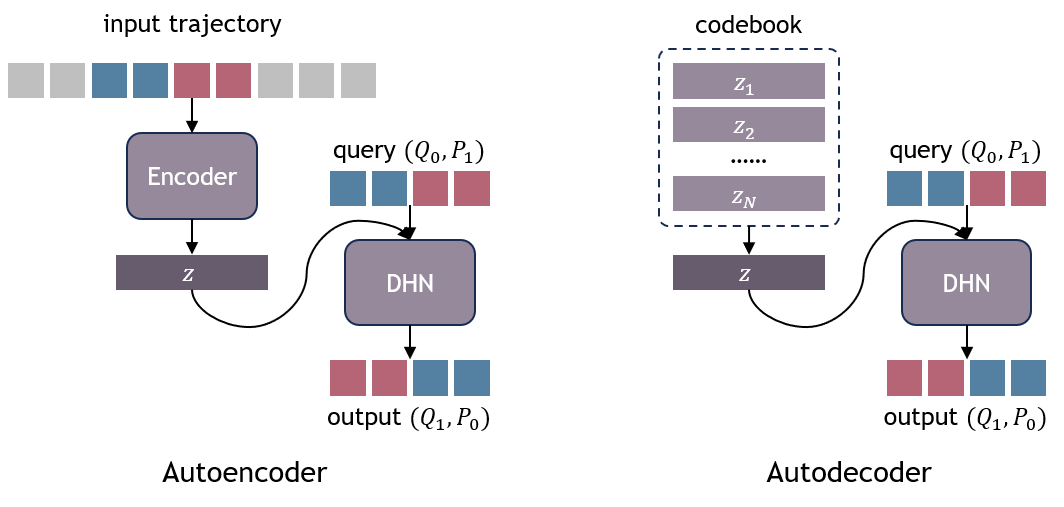}
    \vspace{-2.8em}
    \caption{\textbf{Autodecoder.}
    Instead of encoding the input trajectory with an encoder, we maintain a codebook for the entire dataset with a learnable latent code for each trajectory. 
    Dark purples (in all shades) indicate trainable modules or variables.
    }
    \label{fig:autodecoder}
\end{figure}

\paragraph{Autodecoding}
Rather than relying on an encoder network to infer the global latent code from the trajectory data, we adopt an autodecoder framework \cite{park2019deepsdf}, maintaining a learnable latent code $z$ for each trajectory (Figure \ref{fig:autodecoder}). This approach allows the model to store and refine system-specific embeddings efficiently without requiring a separate encoding process. During training, we jointly optimize the network weights and the codebook. After training, given a novel trajectory, we freeze the network weights and only optimize the latent code for the new trajectory.

\section{Experiments}

We evaluate our model with two settings: the \textit{single pendulum} and the \textit{double pendulum}. Both settings comprise a dataset of simulated trajectories.
The single pendulum is a periodic system where the total energy at each state can be directly computed from $(q,p)$, and thus we use it to evaluate the models' energy conservation ability.
The double pendulum is a chaotic system where small perturbations can lead to diverged future states. 

Unlike prior works \cite{toth2020hamiltonian} which generated data using a fixed set of system parameters while varying initial conditions, we introduce variation by altering the string lengths of the pendulums while keeping initial states fixed (Appendix Figure \ref{fig:exp_setup}).
This modification evaluates whether models can generalize to a broader class of parameterized dynamical systems rather than fitting to a single-instance system.
For both settings, we split the dataset into 1000 training trajectories and 200 testing trajectories. Each trajectory is discretized into 128 time steps.
More details can be found in Appendix \ref{sec:appen:exp_details}.

We test our model with three different tasks corresponding to the three different masking patterns in Figure \ref{fig:inference_types}. They are
(i) next-state prediction (autoregression) for forward simulation,
(ii) representation learning with random masking for physical parameter inference,
and (iii) progressive super-resolution for trajectory interpolation.
These tasks highlight DHN’s adaptability to diverse physical reasoning challenges, testing its ability to generate, infer, and interpolate system dynamics under varying observational constraints.

\subsection{Forward Simulation}

We start with the forward simulation task, where the model predicts the future states of a physical system step-by-step given the initial conditions. 
We implement this by applying a masking strategy within each DHN block, where the last few tokens are masked during training, requiring the model to iteratively refine and denoise them (Figure \ref{fig:inference_types} top).
For one DHN block of block size $b$ and stride $s$, the mask is applied to the last $b-s$ tokens.
At inference time, given the known states at time steps $[0,\cdots,t]$, we apply the DHN block to the time steps $[t-b+1,\cdots,t+s]$, where we use the known states $[t-b+1,\cdots,t]$ to predict the unknown states $[t+1,\cdots,t+s]$.
We experiment with block sizes $b=2,4,8$ with strides $s=b/2$.

\begin{figure*}[t]
    \centering
    \includegraphics[width=\linewidth]{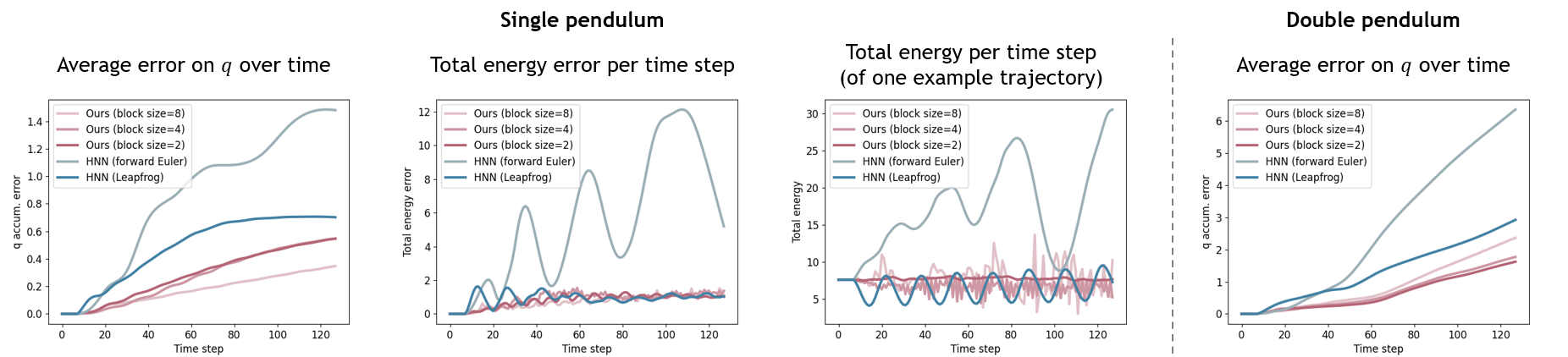}
    \vspace{-2.3em}
    \caption{\textbf{Forward modeling: fitting known trajectories.}
    The results of our method are shown in pink, and the results of HNN with different numerical integrators are shown in different shades of blue.
    1st column: Average state prediction error for the single pendulum.
    2nd column: The total energy for the single pendulum system can be easily calculated with state $(q_t, p_t)$ at each time step analytically. We compare the total energy on the network-predicted states and the ground truth states at each time step.
    3nd column: Predicted total energy over time steps on one example trajectory.
    4th column: Average state prediction error for the double pendulum.
    }
    \label{fig:exp_ar}
    \vspace{-.5em}
\end{figure*}

\paragraph{Fitting known trajectories}
We first evaluate the model's capability to represent known physical trajectories with forward modeling.
In this experiment, we train the model to fit 1000 training trajectories, and we test by giving the first 8 time steps of each trajectory and using the model to predict the future $120$ steps.
As all models are only trained with states of nearby time steps (pairs of adjacent time steps for the baselines, and blocks of $b+s$ states for DHN), small fitting errors can accumulate over time in forward modeling.
Beyond accumulated prediction errors inherent to the network, inaccuracies also arise from numerical integration approximations, which can amplify deviations over time.

Figure \ref{fig:exp_ar} shows the results of our model with different block sizes, compared to HNN \cite{toth2020hamiltonian} with different numerical integrators.
Left and right are the mean squared error (MSE) on the $q$ predictions at each time step for the single and double pendulum systems, respectively.
The middle plots show the averaged total energy error and the evolution of total energy on one example trajectory.
Although HNN is a symplectic network with guaranteed energy conservation, the numerical integrator can still induce uncontrollable energy drifts.
This additional numerical error is particularly inevitable with forward methods. 
While this can be addressed by variational integration methods with implicit state optimizations, the convergence of optimization relies on the knowledge of all possible states including the ones not on the trajectory, which greatly increases the data consumption for training the network.
For our DHN, the state optimization per time step is modeled by the denoising mechanism without the need for a variational integrator.
With block size 2, our model conserves the total energy stably. Increased block sizes can cause energy fluctuations at long time ranges, but this fluctuation doesn't show an obvious inclination of energy drift.

\begin{figure*}[t]
    \centering
    \includegraphics[width=\linewidth]{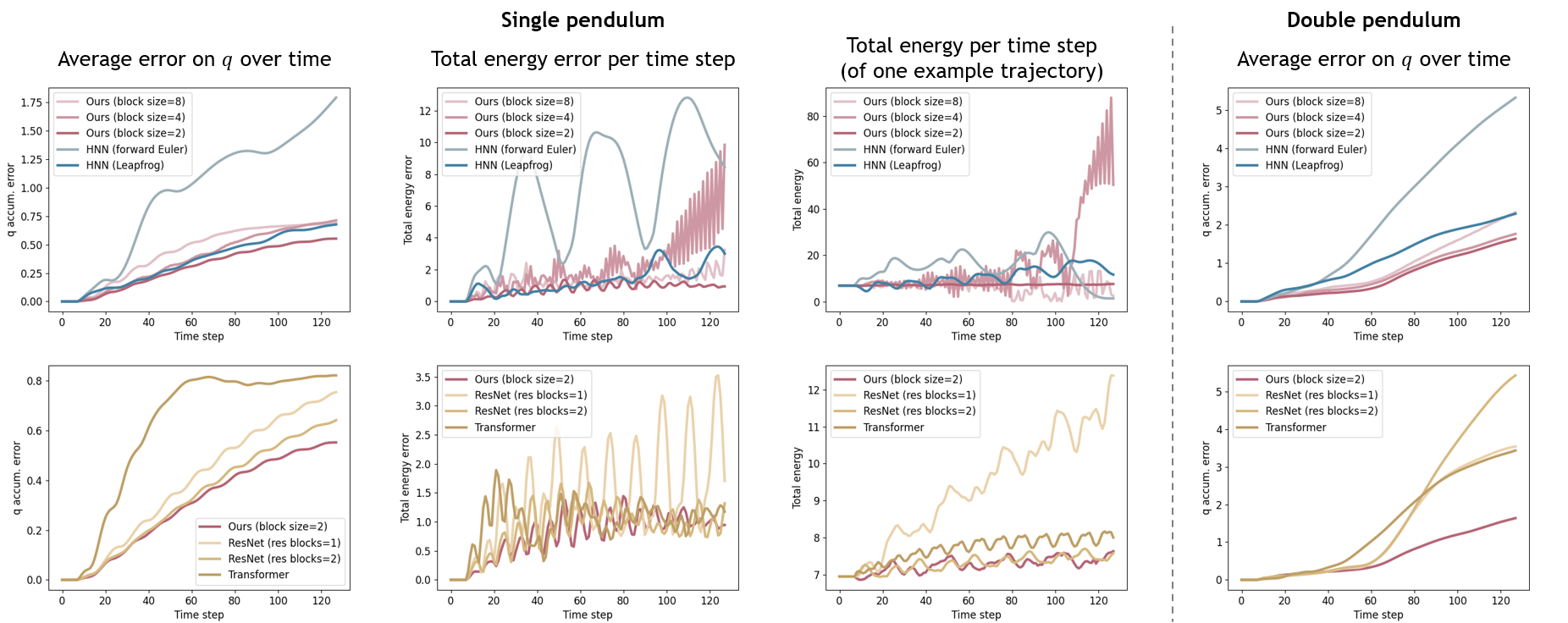}
    \vspace{-2.3em}
    \caption{\textbf{Forward modeling: completion on novel trajectories.}
    Top row: Comparison between our method (shown in pink) and HNN with different numerical integrators (shown in blue).
    Bottom row: Comparison between our method (shown in pink) and vanilla networks with different architectures (shown in yellow). The vanilla networks directly predict the next state $(q_{t+1}, p_{t+1})$ from the current state $(q_t, p_t)$ with one feedforward step.
    Note that the y-axis scales between the two rows are different.
    }
    \label{fig:exp_ar_partial}
    \vspace{-1em}
\end{figure*}

\paragraph{Completion on novel trajectories}
We then evaluate our models on novel trajectories with partial observations.
Concretely, we give the first 16 time steps in each testing trajectory and use them to optimize for the per-trajectory global latent codes with the network weights frozen, as described in Sec. \ref{sec:network_architecture}. 
After optimizing these latent codes, we use them to predict the next 112 time steps. This task evaluates DHN’s ability to infer system dynamics from sparse initial observations and accurately forecast future states.

Figure \ref{fig:exp_ar_partial} shows our results compared to HNN (top row) and various baseline models without physical constraints (bottom row).
Our DHN with small block sizes shows more accurate state prediction with better energy conservation compared to both baselines.
Large block sizes can cause error explosion at long time ranges as it is hard for our simple 2-layer network to fit very complex multi-state relations.

\subsection{Representation Learning}
\label{sec:repn_learning}

Next, we test the model's ability to effectively encode and distinguish the parameters of different physical systems.
Denoising and random masking are well-established techniques in self-supervised learning, producing state-of-the-art representations in language modeling \cite{devlin2018bert} and vision \cite{vincent2008extracting, he2022masked}.
Here, we apply the random masking pattern (Figure \ref{fig:inference_types} bottom) and study whether similar paradigms can enhance representation learning in dynamic physical systems.

To quantify the quality of the learned representations, we follow the widely adopted self-supervised representation learning paradigm in computer vision \cite{chen2020simple, oord2018representation, he2020momentum, kolesnikov2019revisiting} with feature pre-training and linear probing.
Specifically, we pre-train the autodecoder alongside the codebook using the training set, then freeze the learned representations and train a simple linear regression layer on top to predict system parameters. This approach assesses whether DHN's latent codes capture meaningful physical properties.
We experiment with the double pendulum system and predict the length ratio $l_2/l_1$ (Appendix Figure \ref{fig:exp_setup}), because this physical quantity is dimensionless and therefore invariant under scale normalizations in data preprocessing.

\begin{figure}[t]
    \centering
    \includegraphics[width=.9\linewidth]{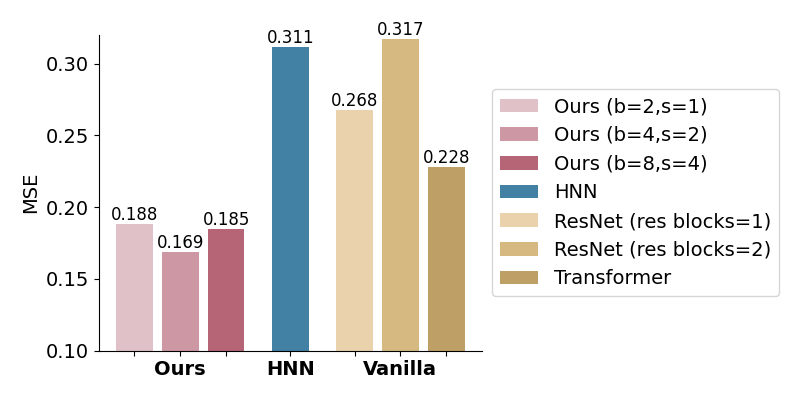}
    \vspace{-1.5em}
    \caption{\textbf{Linear probing on latent codes (MSE $\downarrow$).} We predict $l_2/l_1$ by applying a linear regression layer to the global latent code.}
    \label{fig:exp_repn}
    \vspace{-1em}
\end{figure}

Figure \ref{fig:exp_repn} shows the linear probing results of our DHN with different block sizes (with $s=b/2$), compared to the HNN and vanilla networks. Our model achieves a much lower MSE compared to the baseline networks.
As illustrated in Figure \ref{fig:blockwise_hnn}, HNN can be viewed as a special case of our Hamiltonian block with kernel size and stride being 1, which is of the most locality. The block sizes and strides we introduce allow the model to observe the system at different scales. In this double pendulum system, a block size of 4 is the best temporal scale for inferring its parameters.

\begin{figure}[t]
    \centering
    \begin{subfigure}[t]{\linewidth}
        \centering
        \includegraphics[width=\linewidth]{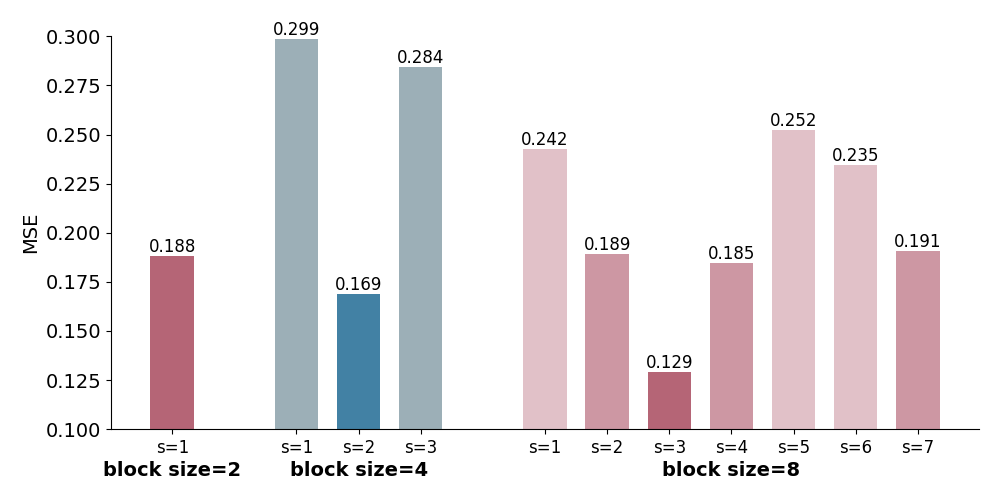}
        \vspace{-1.8em}
        \caption{Results for different block sizes and strides (MSE $\downarrow$).
        Appropriate input-output overlaps with block size $b$ and stride $s$ around $s \approx b/2$ lead to better results.
        }
        \label{fig:exp_repn_ablation:results}
    \end{subfigure}
    \begin{subfigure}[t]{\linewidth}
        \centering
        \includegraphics[width=.8\linewidth]{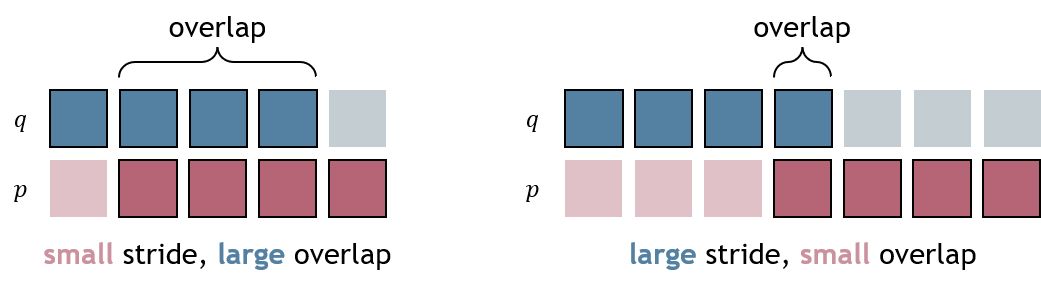}
        \vspace{-.5em}
        \caption{The overlaps between network inputs and outputs induced by different block sizes and strides.}
        \label{fig:exp_repn_ablation:explain}
    \end{subfigure}
    \vspace{-1em}
    \caption{\textbf{Linear probing for different DHN parameters.}}
    \label{fig:exp_repn_ablation}
\end{figure}

Figure \ref{fig:exp_repn_ablation} shows the results of DHN with different block sizes and strides.
As in \ref{fig:exp_repn_ablation:explain}, the input and output states of a Hamiltonian block have an overlapped region of $b-s$ time steps.
The generalized energy conservation of the Hamiltonian block relies on the overlapped region having identical inputs and outputs. During training, this constraint is imposed on the network as part of the state prediction loss.
A larger overlap imposes stronger regularizations on the network, but encourages the network to enforce more of this self-coherence constraint instead of more inter-state relations.
Conversely, reducing overlap while increasing stride encourages the model to incorporate information from more temporally distant states, but at the cost of weaker self-coherence constraints, which can impact stability.
In the extreme case where the overlap equals the block size $b$ and the stride is zero, the DHN block has identical inputs and outputs and the training loss degenerates to the self-coherence constraint.
HNN is another special case with zero overlap (because block size is 1, overlap can only be zero).
As shown in \ref{fig:exp_repn_ablation:explain}, for our simple two-layer transformer, the best block sizes and strides are around $s \approx b/2$ with a moderate amount of overlap.

\subsection{Trajectory Interpolation}

\begin{figure}[t]
    \centering
    \includegraphics[width=.95\linewidth]{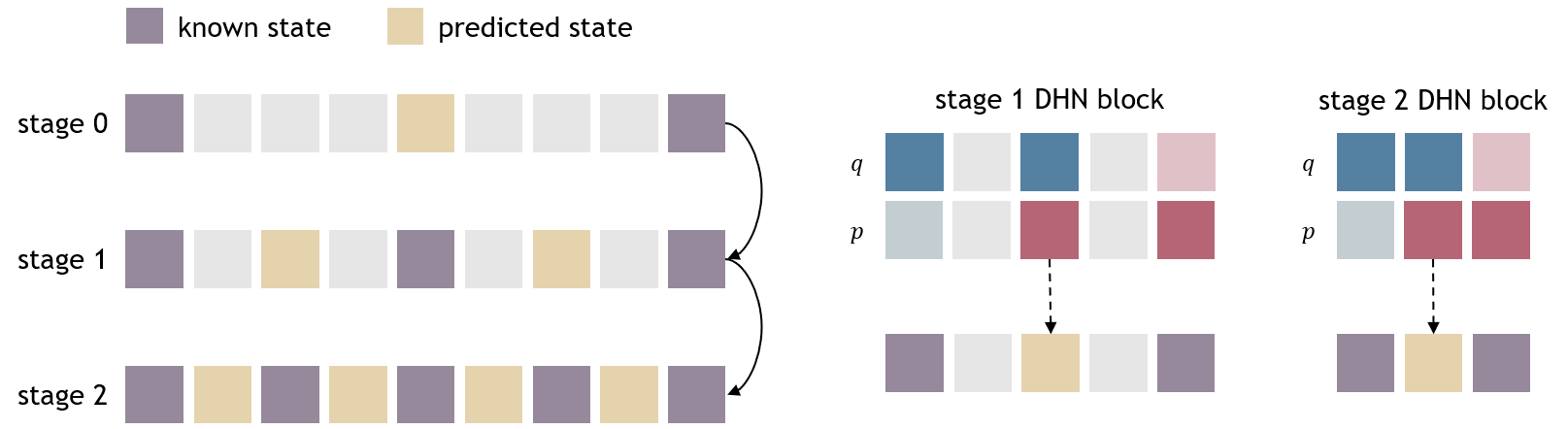}
    \vspace{-1em}
    \caption{\textbf{Interpolation as progressive super-resolution.}
    Left: The three stages for 2$\times$ super-resolution repeated twice.
    Right: DHN blocks for different stages of different sparsity.}
    \label{fig:progressive_superres}
    \vspace{-.5em}
\end{figure}

To demonstrate the flexibility of the DHN block, we show trajectory interpolation (super-resolution) with the masking pattern in Figure \ref{fig:inference_types} middle.
We conduct 4$\times$ super-resolution by repeatedly applying 2$\times$ super-resolutions.
As shown in Figure \ref{fig:progressive_superres} left.
We construct a DHN block with $b=2, s=1$ for each stage. The blocks for trajectories of different sparsity are shown in Figure \ref{fig:progressive_superres} right. The mask is applied to the middle state and the two states at the side are known.

Each trajectory is associated with a shared global latent code across all three super-resolution stages, forming a structured codebook for the training set. During training, both the network weights and these latent codes are optimized jointly across the progressive refinement stages (0, 1, 2).
At inference time, given a novel trajectory with known states only at the sparsest level (stage 0), we freeze all network weights in the DHN blocks and optimize for the global latent code with stage 0. After this test-time optimization (autodecoding), we apply the stage-1, 2 DHN blocks to progressively denoise the unknown states in between the known states.

We evaluate the models with two test settings: (i) trajectories with the same initial states as the training ones, and (ii) trajectories of unseen initial states. To set this up, we crop all training trajectories to time steps $[0,\cdots, 64]$. For each trajectory in the test set, we divide it into two segments: time steps $[0,\cdots,64]$ and $[65,\cdots,128]$, the former having the same initial state as the training set and the latter having different initial states.

\begin{figure}[t]
    \vspace{-.5em}
    \centering
    \includegraphics[width=.87\linewidth]{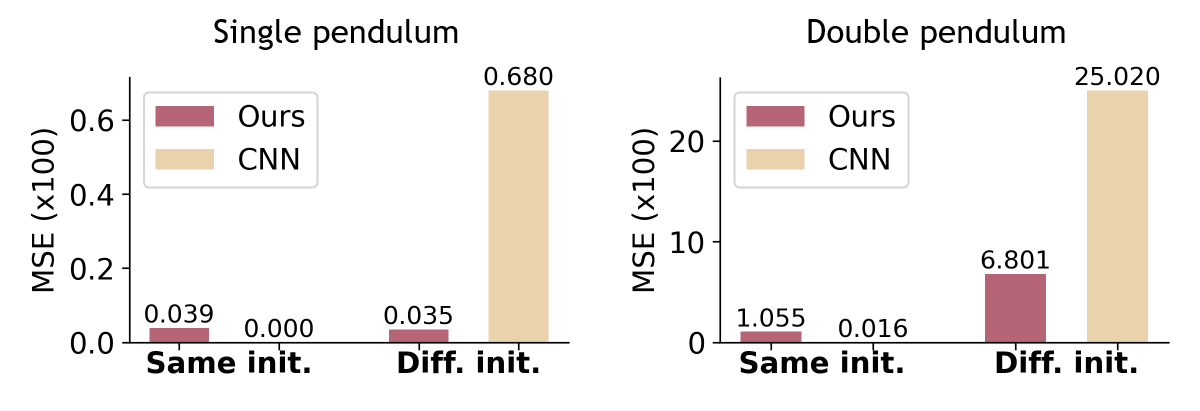}
    \vspace{-1.5em}
    \caption{\textbf{Interpolation (super-resolution) results (MSE$\downarrow$).} We compare the performance of DHN ({\it Ours}) to a CNN-based implementation ({\it CNN}). All MSE values are scaled by 100 for improved precision in decimal representation in the plots.}
    \label{fig:exp_superres}
    \vspace{-1em}
\end{figure}

We compare our model to a Convolutional Neural Network (CNN) for super-resolution.
Figure \ref{fig:exp_superres} shows our results. For the trajectories with the same initial state as the training data, both models show good interpolation results with lower MSEs. The baseline CNN shows slightly better results, as it has no regularization in itself and can easily overfit the training trajectories.
For testing trajectories with unseen initial states, the CNN struggles to generalize, as its interpolations rely heavily on the training distribution. In contrast, DHN demonstrates strong generalization, as its physically constrained representations enable it to infer plausible intermediate states even under distribution shifts.
\section{Discussions and Conclusion}
\label{sec:conclusions}
Balancing flexibility with physical constraints is crucial for advancing physics-based learning. Just as unified architectures in NLP and vision (e.g., transformers) adapt to diverse tasks while maintaining core inductive biases, we explore whether a single model can handle tasks ranging from global parameter inference to local state relations, without sacrificing physical consistency.

A key question that we examined is: {\it What defines physical reasoning in deep learning?}
Beyond next-state prediction, it encompasses parameter estimation, system identification, and discovering high-level relationships in dynamical systems. We envision physics-based learning evolving toward adaptable frameworks that fluidly transition across tasks while maintaining physical rigor.

Another core concept that we reconsidered is: {\it What is physical simulation?}  
Simulation is traditionally framed as a sequential process, where trajectories unfold step by step from an initial state. We reformulate it as a global, temporally consistent reconstruction, taking inspirations from recent video generative models that denoise full sequences rather than predicting frame-by-frame \cite{chi2023diffusion}.

We also studied: {\it What physical attributes should a neural network possess?} While PDE-based methods impose local constraints, our findings suggest that key physical properties can emerge through data-driven learning, much like vision models infer semantics without explicit object detectors.

While our current work provides increased flexibility in Hamiltonian-based network designs, we recognize certain limitations.  One key limitation is computational cost: Our model requires more intensive gradient computations than baseline transformers.
In addition, current experiments focus on small-scale systems with simple temporal dynamics; scaling to complex spatial-temporal systems may benefit from hierarchical or attention-based architectures inspired by modern vision models.

We believe that physics-based learning is on the verge of a major transformation, similar to the rise of self-supervised learning in vision and NLP.
By reframing physical reasoning as a reconstruction problem—predicting system states from partial or corrupted inputs—we move toward a unified modeling paradigm that blends deep learning flexibility with the rigor of physical laws.

\section*{Impact Statement}

This work aims to advance scientific studies by developing AI tools for physics-based reasoning. By incorporating physical constraints into neural networks, we seek to improve the explainability and reliability of learning-based models for scientific applications. However, as with other machine learning approaches, applying neural networks to scientific problems requires caution. Neural networks can exhibit hallucinations or spurious correlations, which may lead to misleading scientific conclusions if not properly validated.

While enforcing physical constraints can enhance trust in AI-driven modeling, it does not eliminate the need for rigorous verification, especially when analyzing experimental data. Users must remain mindful of the limitations of learned representations and ensure that conclusions drawn from AI-assisted analyses are supported by physical principles and empirical validation.

\section*{Acknowledgements}
We thank Rell the cat for her photo in Figure \ref{fig:teaser}.
We also thank Tianwei Yin, Tianyuan Zhang, Shivam Duggal, Yichen Li, Carolina Cuesta-Lázaro, and Katherine L. Bouman for their helpful discussions.
C. Deng and L. Guibas are in part supported by the Toyota Research Institute University 2.0 Program and a Vannevar Bush Faculty Fellowship.
B. Y. Feng and W. T. Freeman are in part supported by the NSF Award 2019786 (The NSF AI Institute for Artificial Intelligence and Fundamental Interactions) and the NSF CIF Award 1955864 (Occlusion and Directional Resolution in Computational Imaging).
C. Garraffo is funded by AstroAI at the Center for Astrophysics at  Harvard \& Smithsonian. A. Garbarz is supported by UBA and CONICET and through the grants PICT 2021-00644, PIP 112202101 00685CO and UBACYT 20020220400140BA. R. Walters is supported by NSF 2134178.


\bibliography{example_paper}

\begin{thebibliography}{25}
\providecommand{\natexlab}[1]{#1}
\providecommand{\url}[1]{\texttt{#1}}
\expandafter\ifx\csname urlstyle\endcsname\relax
  \providecommand{\doi}[1]{doi: #1}\else
  \providecommand{\doi}{doi: \begingroup \urlstyle{rm}\Url}\fi

\bibitem[Chen et~al.(2020)Chen, Kornblith, Norouzi, and Hinton]{chen2020simple}
Chen, T., Kornblith, S., Norouzi, M., and Hinton, G.
\newblock A simple framework for contrastive learning of visual representations.
\newblock In \emph{International conference on machine learning}, pp.\  1597--1607. PMLR, 2020.

\bibitem[Chen et~al.(2018)Chen, Rubanova, Bettencourt, and Duvenaud]{chen2018neural}
Chen, T.~Q., Rubanova, Y., Bettencourt, J., and Duvenaud, D.~K.
\newblock Neural ordinary differential equations.
\newblock In \emph{Neural Information Processing Systems}, 2018.
\newblock URL \url{https://api.semanticscholar.org/CorpusID:49310446}.

\bibitem[Chi et~al.(2023)Chi, Xu, Feng, Cousineau, Du, Burchfiel, Tedrake, and Song]{chi2023diffusion}
Chi, C., Xu, Z., Feng, S., Cousineau, E., Du, Y., Burchfiel, B., Tedrake, R., and Song, S.
\newblock Diffusion policy: Visuomotor policy learning via action diffusion.
\newblock \emph{The International Journal of Robotics Research}, pp.\  02783649241273668, 2023.

\bibitem[Cranmer et~al.(2020)Cranmer, Greydanus, Hoyer, Battaglia, Spergel, and Ho]{cranmer2020lagrangian}
Cranmer, M., Greydanus, S., Hoyer, S., Battaglia, P.~W., Spergel, D.~N., and Ho, S.
\newblock Lagrangian neural networks.
\newblock \emph{ArXiv}, abs/2003.04630, 2020.
\newblock URL \url{https://api.semanticscholar.org/CorpusID:212644628}.

\bibitem[Devlin(2018)]{devlin2018bert}
Devlin, J.
\newblock Bert: Pre-training of deep bidirectional transformers for language understanding.
\newblock \emph{arXiv preprint arXiv:1810.04805}, 2018.

\bibitem[Dupont et~al.(2019)Dupont, Doucet, and Teh]{dupont2019augmented}
Dupont, E., Doucet, A., and Teh, Y.~W.
\newblock Augmented neural odes.
\newblock \emph{ArXiv}, abs/1904.01681, 2019.
\newblock URL \url{https://api.semanticscholar.org/CorpusID:102487914}.

\bibitem[Finzi et~al.(2020)Finzi, Wang, and Wilson]{finzi2020simplifying}
Finzi, M., Wang, K.~A., and Wilson, A.~G.
\newblock Simplifying hamiltonian and lagrangian neural networks via explicit constraints.
\newblock \emph{ArXiv}, abs/2010.13581, 2020.
\newblock URL \url{https://api.semanticscholar.org/CorpusID:225067856}.

\bibitem[Gonzalez(1996)]{gonzalez1996time}
Gonzalez, O.
\newblock Time integration and discrete hamiltonian systems.
\newblock \emph{Journal of Nonlinear Science}, 6:\penalty0 449--467, 1996.

\bibitem[Greydanus et~al.(2019)Greydanus, Dzamba, and Yosinski]{greydanus2019hamiltonian}
Greydanus, S., Dzamba, M., and Yosinski, J.
\newblock Hamiltonian neural networks.
\newblock In \emph{Neural Information Processing Systems}, 2019.
\newblock URL \url{https://api.semanticscholar.org/CorpusID:174797937}.

\bibitem[He et~al.(2020)He, Fan, Wu, Xie, and Girshick]{he2020momentum}
He, K., Fan, H., Wu, Y., Xie, S., and Girshick, R.
\newblock Momentum contrast for unsupervised visual representation learning.
\newblock In \emph{Proceedings of the IEEE/CVF conference on computer vision and pattern recognition}, pp.\  9729--9738, 2020.

\bibitem[He et~al.(2022)He, Chen, Xie, Li, Doll{\'a}r, and Girshick]{he2022masked}
He, K., Chen, X., Xie, S., Li, Y., Doll{\'a}r, P., and Girshick, R.
\newblock Masked autoencoders are scalable vision learners.
\newblock In \emph{Proceedings of the IEEE/CVF conference on computer vision and pattern recognition}, pp.\  16000--16009, 2022.

\bibitem[Ho et~al.(2020)Ho, Jain, and Abbeel]{ho2020denoising}
Ho, J., Jain, A., and Abbeel, P.
\newblock Denoising diffusion probabilistic models.
\newblock \emph{Advances in neural information processing systems}, 33:\penalty0 6840--6851, 2020.

\bibitem[Jin et~al.(2024)Jin, Jiang, Tan, Zhang, Bi, Zhang, Luan, Snavely, and Xu]{jin2024lvsm}
Jin, H., Jiang, H., Tan, H., Zhang, K., Bi, S., Zhang, T., Luan, F., Snavely, N., and Xu, Z.
\newblock Lvsm: A large view synthesis model with minimal 3d inductive bias.
\newblock \emph{arXiv preprint arXiv:2410.17242}, 2024.

\bibitem[Kolesnikov et~al.(2019)Kolesnikov, Zhai, and Beyer]{kolesnikov2019revisiting}
Kolesnikov, A., Zhai, X., and Beyer, L.
\newblock Revisiting self-supervised visual representation learning.
\newblock In \emph{Proceedings of the IEEE/CVF conference on computer vision and pattern recognition}, pp.\  1920--1929, 2019.

\bibitem[Li et~al.(2020)Li, Kovachki, Azizzadenesheli, Liu, Bhattacharya, Stuart, and Anandkumar]{li2021fourier}
Li, Z.-Y., Kovachki, N.~B., Azizzadenesheli, K., Liu, B., Bhattacharya, K., Stuart, A.~M., and Anandkumar, A.
\newblock Fourier neural operator for parametric partial differential equations.
\newblock \emph{ArXiv}, abs/2010.08895, 2020.
\newblock URL \url{https://api.semanticscholar.org/CorpusID:224705257}.

\bibitem[Ljung(1999)]{ljung1998system}
Ljung, L.
\newblock System identification: theory for the user.
\newblock 1999.
\newblock URL \url{https://api.semanticscholar.org/CorpusID:53821855}.

\bibitem[Oord et~al.(2018)Oord, Li, and Vinyals]{oord2018representation}
Oord, A. v.~d., Li, Y., and Vinyals, O.
\newblock Representation learning with contrastive predictive coding.
\newblock \emph{arXiv preprint arXiv:1807.03748}, 2018.

\bibitem[Park et~al.(2019)Park, Florence, Straub, Newcombe, and Lovegrove]{park2019deepsdf}
Park, J.~J., Florence, P., Straub, J., Newcombe, R., and Lovegrove, S.
\newblock Deepsdf: Learning continuous signed distance functions for shape representation.
\newblock In \emph{Proceedings of the IEEE/CVF conference on computer vision and pattern recognition}, pp.\  165--174, 2019.

\bibitem[Radford et~al.(2019)Radford, Wu, Child, Luan, Amodei, Sutskever, et~al.]{radford2019language}
Radford, A., Wu, J., Child, R., Luan, D., Amodei, D., Sutskever, I., et~al.
\newblock Language models are unsupervised multitask learners.
\newblock \emph{OpenAI blog}, 1\penalty0 (8):\penalty0 9, 2019.

\bibitem[Raissi et~al.(2019)Raissi, Perdikaris, and Karniadakis]{raissi2019physics}
Raissi, M., Perdikaris, P., and Karniadakis, G.~E.
\newblock Physics-informed neural networks: A deep learning framework for solving forward and inverse problems involving nonlinear partial differential equations.
\newblock \emph{J. Comput. Phys.}, 378:\penalty0 686--707, 2019.
\newblock URL \url{https://api.semanticscholar.org/CorpusID:57379996}.

\bibitem[Song et~al.(2020)Song, Sohl-Dickstein, Kingma, Kumar, Ermon, and Poole]{song2020score}
Song, Y., Sohl-Dickstein, J., Kingma, D.~P., Kumar, A., Ermon, S., and Poole, B.
\newblock Score-based generative modeling through stochastic differential equations.
\newblock \emph{arXiv preprint arXiv:2011.13456}, 2020.

\bibitem[Toth et~al.(2019)Toth, Rezende, Jaegle, Racani{\`e}re, Botev, and Higgins]{toth2020hamiltonian}
Toth, P., Rezende, D.~J., Jaegle, A., Racani{\`e}re, S., Botev, A., and Higgins, I.
\newblock Hamiltonian generative networks.
\newblock \emph{ArXiv}, abs/1909.13789, 2019.
\newblock URL \url{https://api.semanticscholar.org/CorpusID:203593936}.

\bibitem[Vincent et~al.(2008)Vincent, Larochelle, Bengio, and Manzagol]{vincent2008extracting}
Vincent, P., Larochelle, H., Bengio, Y., and Manzagol, P.-A.
\newblock Extracting and composing robust features with denoising autoencoders.
\newblock In \emph{Proceedings of the 25th international conference on Machine learning}, pp.\  1096--1103, 2008.

\bibitem[Zhong et~al.(2019)Zhong, Dey, and Chakraborty]{zhong2020symplectic}
Zhong, Y.~D., Dey, B., and Chakraborty, A.
\newblock Symplectic ode-net: Learning hamiltonian dynamics with control.
\newblock \emph{ArXiv}, abs/1909.12077, 2019.
\newblock URL \url{https://api.semanticscholar.org/CorpusID:202889233}.

\bibitem[Zhong et~al.(2020)Zhong, Dey, and Chakraborty]{zhong2020dissipative}
Zhong, Y.~D., Dey, B., and Chakraborty, A.
\newblock Dissipative symoden: Encoding hamiltonian dynamics with dissipation and control into deep learning.
\newblock \emph{ArXiv}, abs/2002.08860, 2020.
\newblock URL \url{https://api.semanticscholar.org/CorpusID:211205165}.

\end{thebibliography}
\bibliographystyle{icml2025}

\newpage
\appendix
\section{Discrete Left Hamiltonian $H^-$}
\label{sec:appen:h_minus}

The discrete right Hamiltonian $H^-$ gives the equation of motion in the form
\begin{align}
    q_t &= -\nabla_p H^-(q_{t+1}, p_t), \label{eq:h_minus_q} \\
    p_{t+1} &= -\nabla_q H^-(q_{t+1}, p_t). \label{eq:h_minus_p}
\end{align}
It can be a first-order approximation of the continuous Hamiltonian $\gH$ by
\begin{align}
    q_t &= q_{t+1} - \Delta t \nabla_p \gH(q_t, p_{t+1}), \label{eq:h_minus_continuous_q} \\
    p_{t+1} &= p_t - \Delta t \nabla_q \gH(q_t, p_{t+1}). \label{eq:h_minus_continuous_p}
\end{align}

When extended blocked states, the block-wise discrete left Hamiltonian is defined as 
\begin{align}
\label{eq:block_h_minus}
    Q_{t}^{t+b} &= -\nabla_P H^-(Q_{t+s}^{t+s+b}, P_{t}^{t+b}), \\
    P_{t+s}^{t+s+b} &= -\nabla_Q H^-(Q_{t+s}^{t+s+b}, P_{t}^{t+b}).
\end{align}

Fig. \ref{fig:left_and_right} below illustrates the relation between discrete left and right Hamiltonians in both classical forms and our block-wise extensions. Both the left and right Hamiltonians take each other's outputs as inputs.

\begin{figure}[h]
    \centering
    \includegraphics[width=\linewidth]{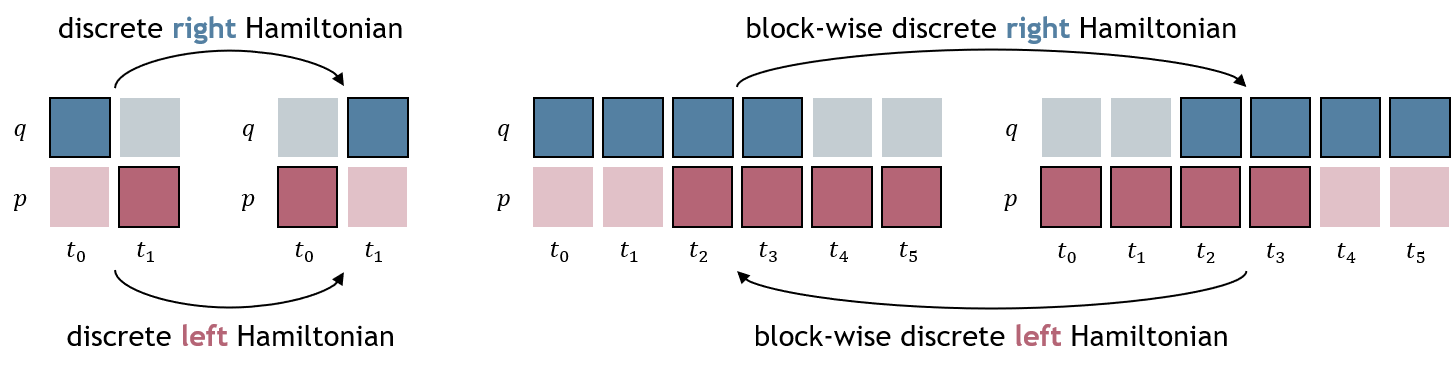}
    \caption{\textbf{Discrete left and right Hamiltonian blocks.} Both of them take each other's outputs as inputs.}
    \label{fig:left_and_right}
\end{figure}

\section{Physical Interpretations for DHN}
\label{sec:appen:block_hnn}

\begin{figure}[h]
    \centering
    \includegraphics[width=\linewidth]{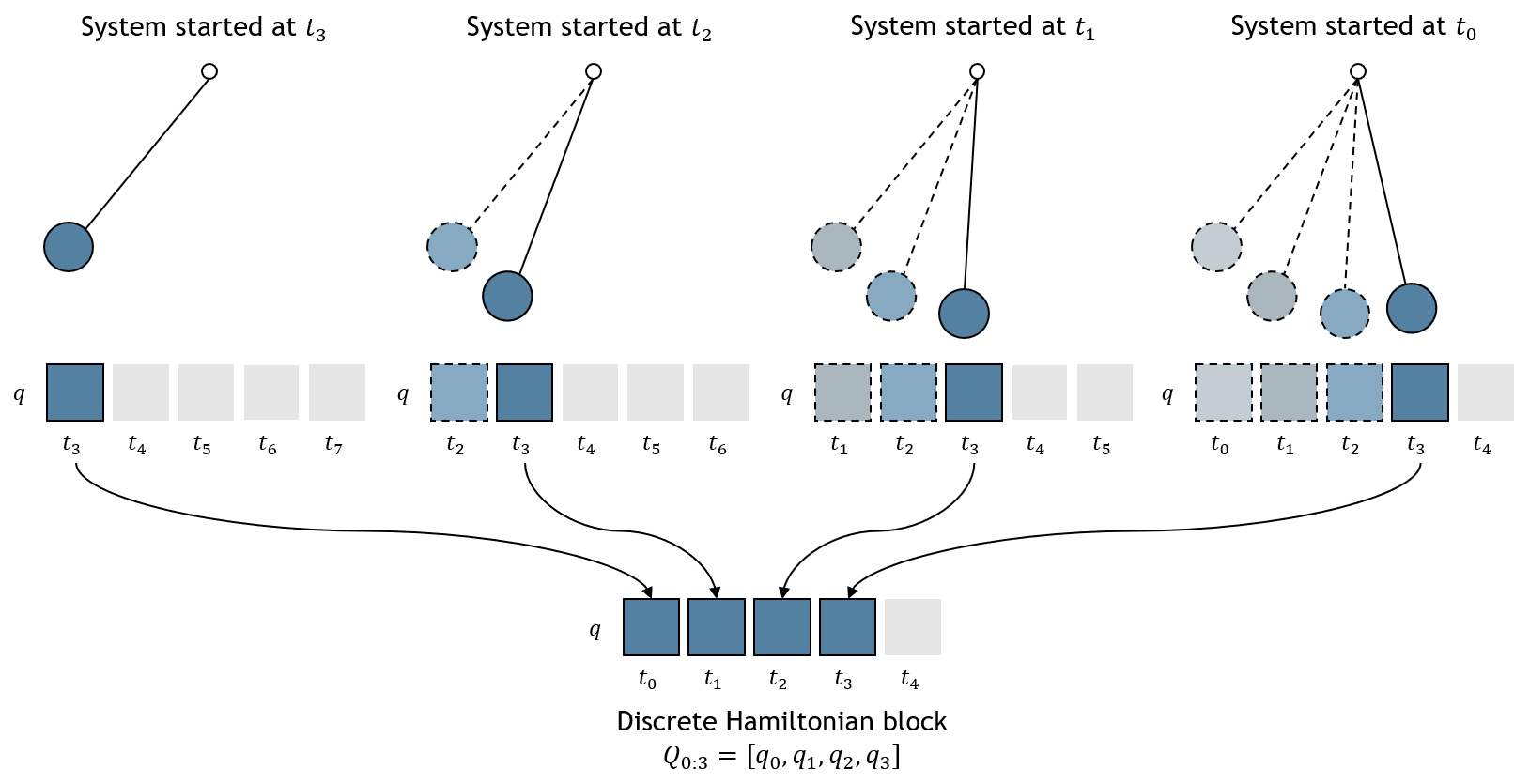}
    \caption{
    \textbf{Physical interpretations of block-wise discrete Hamiltonian.}
    The states within a discrete Hamiltonian block can be interpreted as those of identical physical systems, each starting at a different time
    }
    \label{fig:block_explain}
\end{figure}

In this section, we discuss whether extending the discrete Hamiltonian to block sizes and strides greater than 1 still allows for explicit physical interpretations. Specifically, we address the following two questions:

(i) \textit{What is the conserved quantity with the block-wise Hamiltonian?}
For a discrete Hamiltonian block of size  $b$, the conserved quantity is the sum of the total energy of $b$ independent states. More specifically, the states within a discrete Hamiltonian block can be interpreted as those of identical physical systems, each starting at a different time. Figure \ref{fig:block_explain} provides an illustration of this concept.

Consider the case where the block size is $b=4$. Suppose we have four identical physical systems, each initialized at different times: $t_0, t_1, t_2, t_3$. By time $t_3$, these systems will have evolved for 0, 1, 2, and 3 time steps, respectively. If we take their states at $t_3$ and stack them together, we obtain a state block that effectively represents four consecutive states spanning four time steps within a single system. Importantly, the four states at $t_3$ remain independent, as the four duplicated systems do not interact with one another.
Thus, the conserved quantity in this framework is the total energy summed across all these identical, non-interacting systems.

(ii) \textit{What are the relaxations compared to classical discrete Hamiltonian?}
When extending the classical discrete Hamiltonian to a block-wise formulation, certain physical constraints are relaxed. The two main relaxations are as follows:

First, instead of conserving the energy of each individual state, the block-wise Hamiltonian conserves the total energy summed over $b$ states. This allows for different energy distributions across the $b$ states, making the constraint weaker than enforcing per-state energy conservation.

Second, as discussed in Sec. \ref{sec:repn_learning}, when the stride $s$ is smaller than the block size $b$, there is an overlap of $b-s$ between network inputs and outputs. In theory, exact energy conservation (in the generalized form) requires that the overlapping states remain identical. However, in practice, this self-consistency loss is rarely minimized to exactly zero. The extent to which it is minimized depends on factors such as network expressivity, architecture, and hyperparameters $b$ and $s$, which in turn affect how well energy conservation is maintained.

Despite these relaxations, the model still enforces a form of physical consistency across the trajectory. Rather than strictly conserving per-state energy, it shifts toward preserving higher-level conserved quantities. This relaxation also opens the door to developing more abstract notions of physical consistency on latent embeddings instead of the raw observed states.

\section{Denoising Inference}
\label{sec:appen:denoising}

\begin{figure}[h]
    \centering
    \begin{subfigure}[t]{\linewidth}
        \centering
        \includegraphics[width=.8\linewidth]{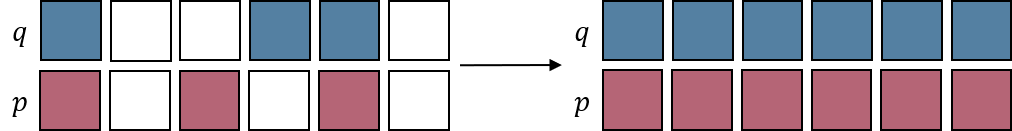}
        \caption{Input and output. DHN blocks of size $b$ and stride $s$ can denoise a stack of $b+s$ states. Colored squares represent known states, while white squares indicate unknown states.}
        \label{fig:denoise_inference_goal}
    \end{subfigure}
    \\\vspace{.5em}
    \begin{subfigure}[t]{\linewidth}
        \centering
        \includegraphics[width=.8\linewidth]{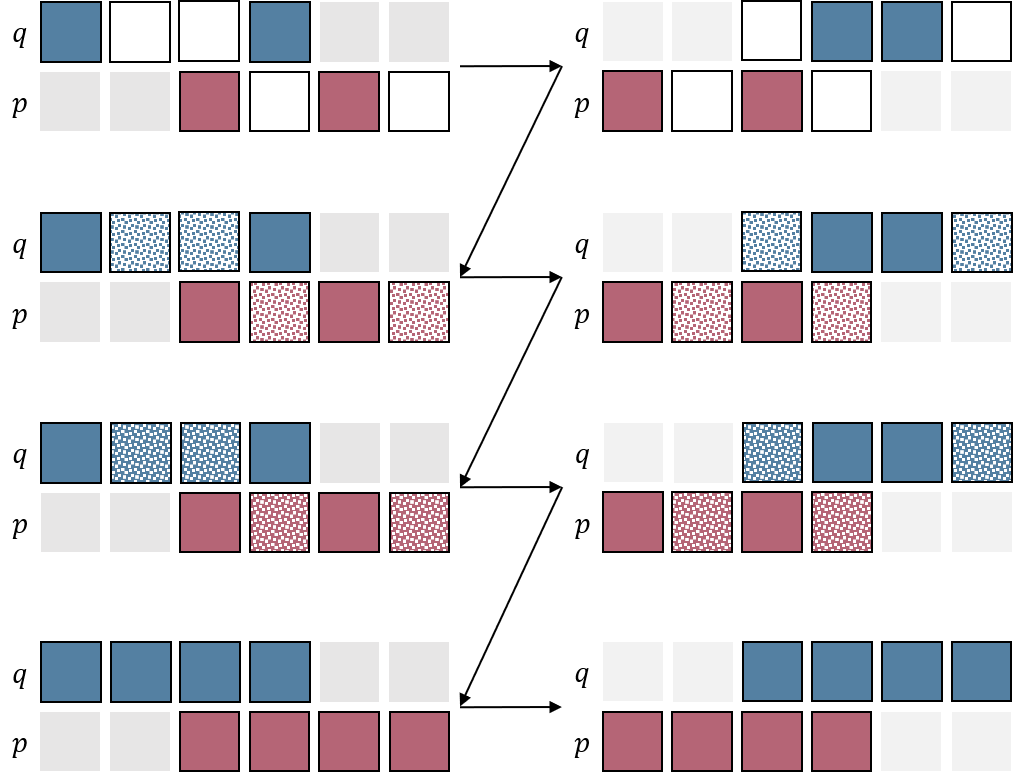}
        \caption{Progressive denoising by iteratively applying block-wise $H^+$ and $H^-$ and gradually decreasing the noise scales.}
        \label{fig:denoise_inference_iterative}
    \end{subfigure}
    \caption{\textbf{Iterative denoising at inference time.}  A pair of DHN blocks, $H^+, H^-$, with block size $b$ and stride $s$, are jointly applied to a stack of $b+s$ states to denoise the unknown blocks.}
    \label{fig:denoise_inference}
\end{figure}

As mentioned in Sec. \ref{sec:denoising_hnn}, unlike training time that applies noises with randomly sampled scales to different unknown states, at inference time, we progressively denoise the unknown states with a sequence of decreasing noise scales that are synchronized on all unknown states.
Fig. \ref{fig:denoise_inference} illustrates the iterative denoising process at inference time with a pair of DHN blocks $H^+$ and $H^-$.

Taking a pair of states $(q_0, p_0)$ as example, given a sequence of noise levels $0=\alpha_0 < \alpha_1 < \cdots < \alpha_N=1$, we begin by sampling $(q_N, p_N)$ from a Gaussian distribution $\gN(0, \rmI)$.
At step $n$, we denoise the states $(q_n, p_n)$ of noise level $\alpha_n$ into $(q_{n-1}, p_{n-1})$ of noise level $\alpha_{n-1}$ via 
\begin{align}
    (\hat{q}_0, \hat{p}_0) &= \texttt{DHN}(q_n, p_n), \\
    (q_{n-1}, p_{n-1}) &= (1-\alpha_{n-1}) (\hat{q}_0, \hat{p}_0) + \alpha_{n-1}\varepsilon,
\end{align}
where $\varepsilon\sim\gN(0, \rmI)$. This is similar to the diffusion models \cite{ho2020denoising, song2020score}.

\section{Experiment Settings}
\label{sec:appen:exp_details}

\begin{figure}[h]
    \centering
    \includegraphics[width=\linewidth]{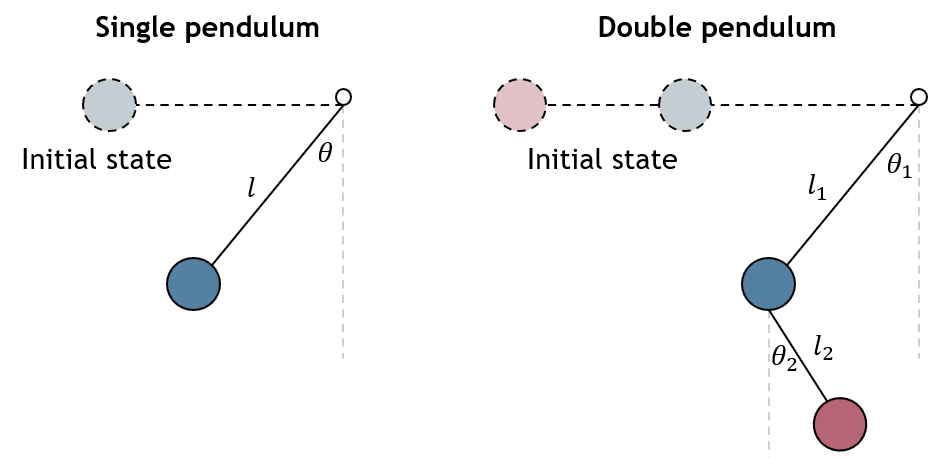}
    \caption{\textbf{Physical systems for the experiments.}
    Circles with dotted lines and swallower colors are the initial states, which are identical to all training and testing trajectories. Circles with solid lines and darker colors illustrate the intermediate states along the simulated trajectory.
    }
    \label{fig:exp_setup}
\end{figure}

Here we elaborate on the details of the two settings we experiment with: the \textit{single pendulum} and the \textit{double pendulum}, as illustrated in Fig. \ref{fig:exp_setup}. In both settings, we first define the generalized coordinate $q$ and the system's Lagrangian $\gL(q, \dot{q})$. The generalized momenta is then defined by $p= \nabla_{\dot{q}}\gL$. We set the gravitational acceleration $g=0.981$.

\paragraph{Single pendulum}
In this system, the varied parameter is the string length $l$, randomly sampled between [0.5, 1.0] for each trajectory. The mass of the ball is set to be $m=1$. The generalized coordinate is defined as $q=\theta$, with initial value $\theta=\pi/2$ for all trajectories.
The Lagrangian of the system is
\begin{equation}
    \gL = \frac{1}{2}m l^2 \dot q^2 -mgl (1-\cos q).
\end{equation}
Here $(q, p)$ are the standard angular position and angular momentum in spherical coordinates.

\paragraph{Double pendulum}
In this system, the varied parameter is the string length $l_2$, randomly sampled between [0.5, 1.5] for each trajectory. The remaining fixed parameters are $l_1=1, m_1=m_2=1$. The generalized coordinate is defined as $q=(\theta_1, \theta_2)$, with initial values $\theta_1=\theta_2=\pi/2$ for all trajectories.
The Lagrangian of the system is
\begin{align}
    \gL =& \frac{1}{2}(m_1+m_2)l_1^2\dot\theta_1^2
    + \frac{1}{2}m_2l_2^2\dot\theta_2^2 \\
    &+ m_2l_1l_2\dot\theta_1\dot\theta_2\cos(\theta_1-\theta_2) \\
    &+ (m_1+m_2)gl_1\cos\theta_1+m_2gl_2\cos\theta_2.
\end{align}

\end{document}